\documentclass[10pt,twocolumn,letterpaper,dvipsnames]{article}
\usepackage[pagenumbers]{cvpr}  

\usepackage[export]{adjustbox}
\usepackage{amsmath,amssymb,amsfonts}  
\usepackage{bm}
\usepackage{csquotes}                  
\usepackage{etoolbox}                  
\usepackage{graphicx}
\usepackage{nicefrac}                  
\usepackage{soul}                      
\usepackage{tikz}                      
\usepackage{textcomp}                  
\usepackage{url}                       
\usepackage{xcolor}                    
\usepackage{xparse}                    

\definecolor{cvprblue}{rgb}{0.21,0.49,0.74}
\usepackage[pagebackref,breaklinks,colorlinks,citecolor=cvprblue]{hyperref}
\hypersetup{
    colorlinks=true,
    linkcolor=cvprblue,
    citecolor=cvprblue,
    urlcolor=cvprblue
}
\usepackage[capitalize]{cleveref}      
\usepackage[accsupp]{axessibility}     
\usepackage{microtype}                 

\newtoggle{nographics}\togglefalse{nographics}

\newtoggle{nocomments}\togglefalse{nocomments}

\begin{document}
    \def\paperID{}
\def\confName{CVPR}
\def\confYear{2026}

\title{CADFS: A Big CAD Program Dataset and Framework for Computer-Aided Design with Large Language Models}

\author{\
    Vladislav Pyatov\textsuperscript{1,2}
    \quad{}Gleb Bobrovskikh\textsuperscript{1}
    \quad{}Saveliy Galochkin\textsuperscript{3,1}
    \quad{}Nikita Boldyrev\textsuperscript{1}
    \\\
    Oleg Voynov\textsuperscript{1,3}
    \quad{}Alexander Filippov\textsuperscript{2}
    \quad{}Gonzalo Ferrer\textsuperscript{1}
    \quad{}Peter Wonka\textsuperscript{4}
    \quad{}Evgeny Burnaev\textsuperscript{1,3}
    \medskip\\\
    \textsuperscript{1}Applied AI Institute
    \quad{}\textsuperscript{2}AI Foundation and Algorithm Lab
    \quad{}\textsuperscript{3}AXXX
    \quad{}\textsuperscript{4}KAUST
}

\newcommand{\todo}[1]{{\color{red}TODO: #1}}

\DeclareGraphicsExtensions{.eps,.pdf,.jpg,.png}
\graphicspath{{src/img/}}
\makeatletter
\iftoggle{nographics}{
    \LetLtxMacro{\includegraphics@orig}{\includegraphics}
    \RenewDocumentCommand{\includegraphics}{ s O{} m }{%
            {\setlength{\fboxsep}{0pt}%
        \colorbox{lightgray}{\phantom{\IfBooleanTF{#1}{\includegraphics@orig*}{\includegraphics@orig}[#2]{#3}}}%
        }%
    }
}{}
\makeatother

\renewcommand{\paragraph}[1]{\vspace{.5em}\noindent\textbf{#1}}  

\twocolumn[{%
    \renewcommand\twocolumn[1][]{#1}%
    \maketitle
    \vspace{-10mm}
    \begin{center}
        \includegraphics[max height=13.2\baselineskip, max width=\textwidth]{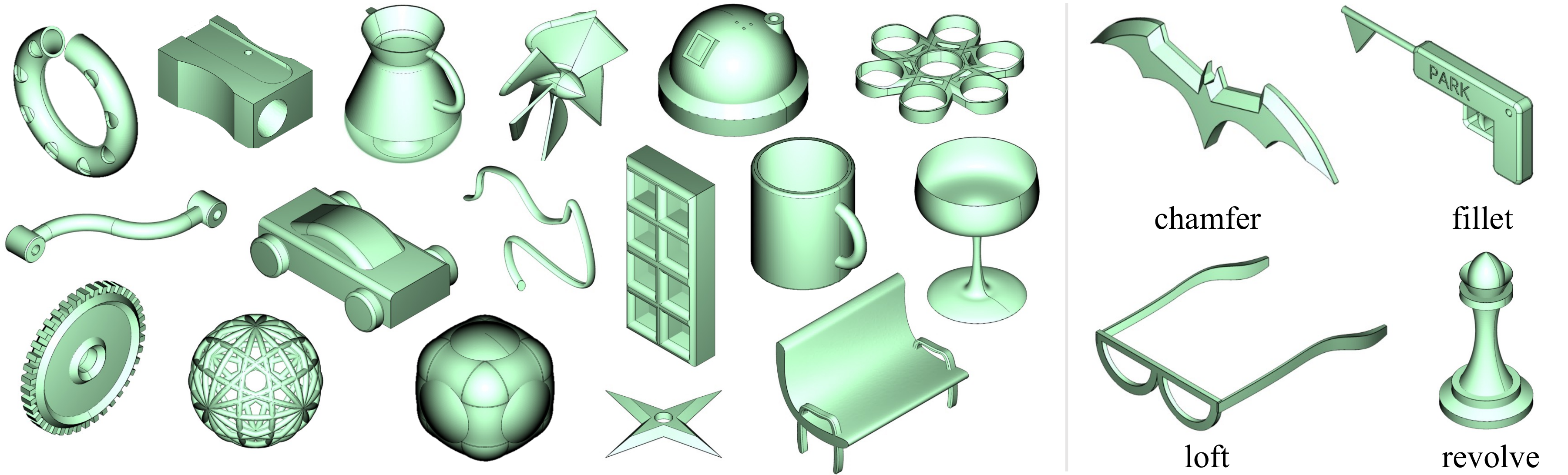}
        \captionof{figure}{
            We propose a new framework for generative CAD with large language models that enables generation of more complex designs compared to prior frameworks.
            Left: designs generated with a VLM based on our framework.
            The core of our framework is a new CAD design history representation that enables a broader range of modeling operations.
            Right: examples using advanced operations.
        }
        \label{fig:teaser}
    \end{center}
}]

\begin{abstract}
We introduce CADFS, a data-centric framework that enables large vision-language models to generate complex CAD design histories.
Existing generative CAD systems are restricted to sketch-extrude operations due to simplified representations and limited datasets.
We address this by introducing a FeatureScript-based representation and constructing a dataset of 450k real-world CAD models spanning 15 modeling operations.
We obtain the dataset via a new pipeline that reconstructs clean, executable FeatureScript programs and provides multimodal annotations.
Fine-tuning a VLM on this representation yields state-of-the-art results in text-conditioned CAD generation and image-based reconstruction, producing more accurate, diverse, and feature-rich designs than prior frameworks.
Ablations show that each individual component of our framework, \ie, the FeatureScript representation, the extended operation set, and representation-aligned textual descriptions, significantly improves performance.
Our framework substantially broadens the complexity and realism achievable in generative CAD.
The CADFS framework and the new dataset are available at \href{https://voyleg.github.io/cadfs/}{voyleg.github.io/cadfs}.
\end{abstract}

\section{Introduction}
\label{sec:intro}

Designing 3D components is a core task across engineering domains such as industrial manufacturing or architecture.
While computer-aided design (CAD) tools have become the standard for assisting this process, creating high-quality 3D models still requires substantial time and expert effort.
Recent advances in AI promise to significantly reduce this effort by enabling the automated creation and reconstruction of CAD models from diverse inputs such as images~\cite{alam2024gencad,li2025caddreamer}, drawings~\cite{chen2024img2cad}, natural language descriptions~\cite{alrashedy2024generating,khan2024text2cad,wang2025texttocad,guan2025cadcoder,xie2025texttocadquery,li2025cadllama}, or point clouds~\cite{dupont2024transcad,ma2024draw,rukhovich2024cadrecode,kolodiazhnyi2025cadrille,xu2024cadmllm,wu2025cmt}.

A growing body of work focuses on generating CAD models in boundary representation (B-rep) format, producing output that is directly suitable for manufacturing pipelines~\cite{xu2024brepgen,wu2025cmt,li2025caddreamer}.
However, practicing engineers typically design geometry not by editing surfaces directly, but by constructing a design history, \ie, a sequence of parametric modeling operations (\eg, sketching curves and extruding them into solids).
This feature-based representation provides greater flexibility and editability, allowing engineers to revise, extend, or refine models throughout the design process.
Consequently, generating CAD models as design histories rather than B-reps is increasingly seen as the more impactful goal for practical computer-aided design automation.

Early approaches to design history generation trained generative models from scratch~\cite{wu2021deepcad,xu2022skexgen,khan2024text2cad,chen2024img2cad}.
Motivated by the rapid advances and strong generalization capabilities of large language and vision-language models (LLMs and VLMs), recent methods adopt these models to generate CAD construction sequences, achieving promising few-shot~\cite{badagabettu2024query2cad,alrashedy2024generating} and fine-tuned results~\cite{xu2024cadmllm,mews2025dont,kolodiazhnyi2025cadrille,guan2025cadcoder}.
Yet, despite these advances, the complexity of designs generated by existing methods remains limited.
A primary bottleneck is the training data: all existing large-scale design history datasets~\cite{wu2021deepcad,khan2024text2cad,xu2024cadmllm,doris2025cadcoder,guan2025cadcoder,rukhovich2024cadrecode,xie2025texttocadquery} contain only designs composed of two basic operations (sketch and extrude).
As a result, the models trained on these datasets struggle to generate parts requiring richer design operations, such as chamfers, fillets, revolves, or lofts.

This limitation stems from the representation of the design history adopted in prior datasets.
In this simplified token-based sequence representation~\cite{wu2021deepcad,khan2024text2cad,xu2024cadmllm}, a new operation in the design history can only refer to the previously issued operations directly.
However, many standard CAD operations (\eg, fillet, chamfer, or loft) act on or reference specific elements of the evolving geometry, such as newly extruded edges.
As a result, this representation inherently limits the available operation set, constraining the expressiveness and complexity of models.
Moreover, its compactness was intended for efficient autoregressive modeling and may not be the best fit for LLMs, required to learn an artificial token syntax from scratch.
Further works adopt more expressive and natural Python-based CAD scripting interfaces~\cite{doris2025cadcoder,guan2025cadcoder,rukhovich2024cadrecode,xie2025texttocadquery} but still restrict the underlying geometry to the same limited operation set.

To address these limitations, we propose \emph{CADFS}, a new data-centric framework for generative CAD with vision-language models.
At the core of CADFS is a new representation of the design history.
To train a generative model, we collect real-world CAD designs created by engineers on the Onshape platform~\cite{onshape}.
In contrast to prior work, we represent the designs in their native form rather than translating them into a custom simplified representation.
This enables us to incorporate a significantly broader range of modeling operations while retaining the full geometric and parametric fidelity of the original designs.
As a result, we use more complex, diverse, and detailed CAD models for training.

We represent each model as an executable program in FeatureScript, Onshape's native language for parametric CAD~\cite{featurescript}.
This code representation is well-suited for LLMs: it is syntactically structured, semantically interpretable, and expresses modeling logic directly rather than through lossy abstractions.
Prior code representations for generative CAD, based on Python wrappers around geometric kernels (\eg, CadQuery), primarily target procedural modeling workflows.
In contrast, FeatureScript is the native language of a CAD system used by practicing engineers.
This makes it directly aligned with real engineering modeling practices.
To the best of our knowledge, we are the first to explore FeatureScript code as a training representation for generative CAD.

To train VLMs to generate designs in this new representation, we introduce a new large-scale dataset for generative CAD.
Since Onshape does not directly expose clean, executable FeatureScript programs for existing models, we develop a data acquisition pipeline to reconstruct high-quality FeatureScript code from Onshape's internal representation.
Our pipeline unifies parameters, removes redundant expressions, resolves implicit references, and produces consistent, human-readable, executable modeling scripts.
To enable multimodal generative design research, we annotate each CAD model with natural language descriptions and provide rendered images and point clouds.
Our dataset includes over 450k real-world CAD designs constructed using 15 common modeling operations, substantially expanding the diversity of feature-based CAD training data.

The VLM trained with this representation achieves state-of-the-art results in text-conditioned CAD generation and image-conditioned CAD reconstruction.
It produces CAD models with greater geometric variety, higher detail, and more accurate structure, while enabling the generation of previously unattainable feature types (see \cref{fig:teaser}).

In summary, we make the following contributions:
\begin{itemize}
    \item We enable learning CAD design history generation with a much broader set of modeling operations (extending from 2 to 15) by introducing a FeatureScript-based representation and collecting a large-scale dataset of real-world CAD designs in this form.
    \item We present the first vision-language model for this representation capable of modeling complex designs beyond sketch and extrude.
    \item Our framework expands the diversity and complexity of generative CAD and advances the state of the art across multiple CAD generation and reconstruction tasks.
\end{itemize}

\section{Related work}
\label{sec:related}

\paragraph{Methods of generative CAD.}
Prior work approaches CAD generation via either neural sequence modeling of construction steps and parameters or program synthesis that produces executable code.
Sequence models trained from scratch optimize next-step likelihood over tokenized histories, sometimes with multimodal conditioning~\cite{wu2021deepcad,xu2022skexgen,xu2023hierarchical,khan2024text2cad,chen2024img2cad,you2025img2cad,ma2024draw,dupont2024transcad}.
More recent program-synthesis methods leverage pre-trained LLM backbones to produce executable scripts, improving robustness and geometric fidelity with execution-in-the-loop verification, or fine-tuning with multimodal representation learning, chain-of-thought planning, and reinforcement-learning~\cite{badagabettu2024query2cad,alrashedy2024generating,mews2025dont,kolodiazhnyi2025cadrille,guan2025cadcoder,wang2025texttocad,li2025cadllama}.
While these works demonstrate the promise of achieving strong results with LLM backbones, they still use a narrow vocabulary of operations limited to sketch and extrude commands.
This affects the complexity and practicality of generated designs and limits generalization to real engineering workflows.
An orthogonal research direction focuses on direct generation of B-reps with diffusion-based methods~\cite{xu2024brepgen,wu2025cmt,li2025caddreamer}, which demonstrate greater diversity and complexity in the generated models.
Yet, these methods do not recover design history and thus lack an important aspect of practical use.

\paragraph{Representations of CAD design history and datasets.}
Early 3D model collections provide either polygonal meshes~\cite{chang2015shapenet,wu2015modelnet,zhou2016thingi10k,mo2019partnet,kim2020largescale} or B-rep geometry~\cite{koch2019abc,cherenkova2020pvdeconv}, but no design histories, limiting their use for modeling construction processes.
Fusion 360 Gallery~\cite{willis2021fusion} and CC3D-Ops~\cite{dupont2022cadopsnet} were the first datasets to tie CAD models to design steps by annotating 3D data with operation labels or sequences.
Yet these datasets are modest in scale and complexity and therefore do not support training large generative models.

Most recent large-scale design history datasets convert models from the Onshape library~\cite{onshape} into either a simplified token-based representation or Python code.
DeepCAD~\cite{wu2021deepcad} standardizes token sequences with quantized parameters.
This enables autoregressive training, but reduces geometric fidelity and forces models to learn an artificial syntax from scratch.
Crucially, this format can only reference prior operations (\eg, sketch, extrude), not the geometric entities that they produce (\eg, edges, faces).
As a result, it cannot express operations such as fillet, loft, or replication that require structured geometric queries, \ie, selecting specific entities that arise during construction.
This restriction limits the diversity and scale of trainable CAD histories.
Derivative datasets~\cite{khan2024text2cad,xu2024cadmllm} expand annotations for conditional generation yet inherit these constraints.

Code-centric datasets~\cite{doris2025cadcoder,guan2025cadcoder,rukhovich2024cadrecode,xie2025texttocadquery} represent design histories as modular, interpretable Python programs, typically based on the CadQuery library~\cite{cadquery}, leveraging pre-trained LLMs' understanding of Python code.
Compared to token sequences, this representation offers richer CAD functionality and supports operations depending on the evolving geometry.
However, it uses indirect or topology-dependent references to the geometric entities (\eg, \textquote{the third edge of the final solid}).
Such referencing is unstable under small modeling edits and fails to preserve the construction history of geometric entities.
This limits the ability of learning-based methods to reliably infer or reproduce complex feature construction steps.
Datasets based on this representation remain limited to primitive operations.

Overall, both representations introduce structural limitations that reduce expressiveness and restrict the available modeling operations.
The corresponding datasets support only sketch and extrude, along with a small set of sketch primitives: lines, circles, and arcs.
Furthermore, conversion from Onshape's native representation requires lossy translation that discards geometric and parametric fidelity, yielding approximations rather than exact reproductions of the originals.
While recent works address different issues (\eg, short sequence length~\cite{li2025mambacad}, or alignment with software UI~\cite{man2025videocad}), the operation vocabulary and design history faithfulness remain the principal bottlenecks.

In contrast, we build our CADFS framework based on Onshape's native representation.
This preserves full geometric and parametric fidelity and supports a substantially broader set of modeling operations, enabling training on more complex, diverse, and detailed CAD histories.
\Cref{tab:datasets} compares our dataset with prior design history datasets.

\begin{table}[t]
    \centering
    \caption{
        Comparison of our dataset with existing design history benchmarks.
        We provide designs in the new representation featuring the largest set of modeling operations and the largest number of real-world designs (CAD-Recode provides synthetic data).
    }
    \begin{tikzpicture}[scale=1.13]
        \node[anchor=south west,inner sep=0] (image) at (0,0) {\includegraphics[max height=8.4\baselineskip, max width=\columnwidth]{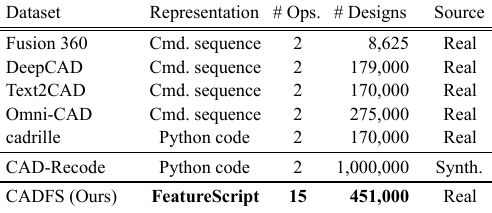}};
        \begin{scope}[shift=(image.north west), x=.988pt, y=-.988pt, anchor=south west, inner sep=0, font={\fontsize{9}{10.6}\selectfont}]
            \node at (1.8, 22.5) {\phantom{Fusion 360}~\cite{willis2021fusion}};
            \node at (1.8, 33) {\phantom{DeepCAD}~\cite{wu2021deepcad}};
            \node at (1.8, 42.5) {\phantom{Text2CAD}~\cite{khan2024text2cad}};
            \node at (1.8, 52.5) {\phantom{Omni-CAD}~\cite{xu2024cadmllm}};
            \node at (1.8, 62.5) {\phantom{cadrille}~\cite{kolodiazhnyi2025cadrille}};
            \node at (1.8, 75) {\phantom{CAD-Recode}~\cite{rukhovich2024cadrecode}};
        \end{scope}
    \end{tikzpicture}
    \label{tab:datasets}
\end{table}

\section{Method}
\label{sec:method}

\begin{figure*}[t]
    \centering
    \vbox to 0pt{\begin{subcaptiongroup}
        \phantomcaption\label{fig:method.pipeline}
        \phantomcaption\label{fig:method.annotation}
        \phantomcaption\label{fig:method.llm}
    \end{subcaptiongroup}}
    \includegraphics[max width=\textwidth,max height=14\baselineskip]{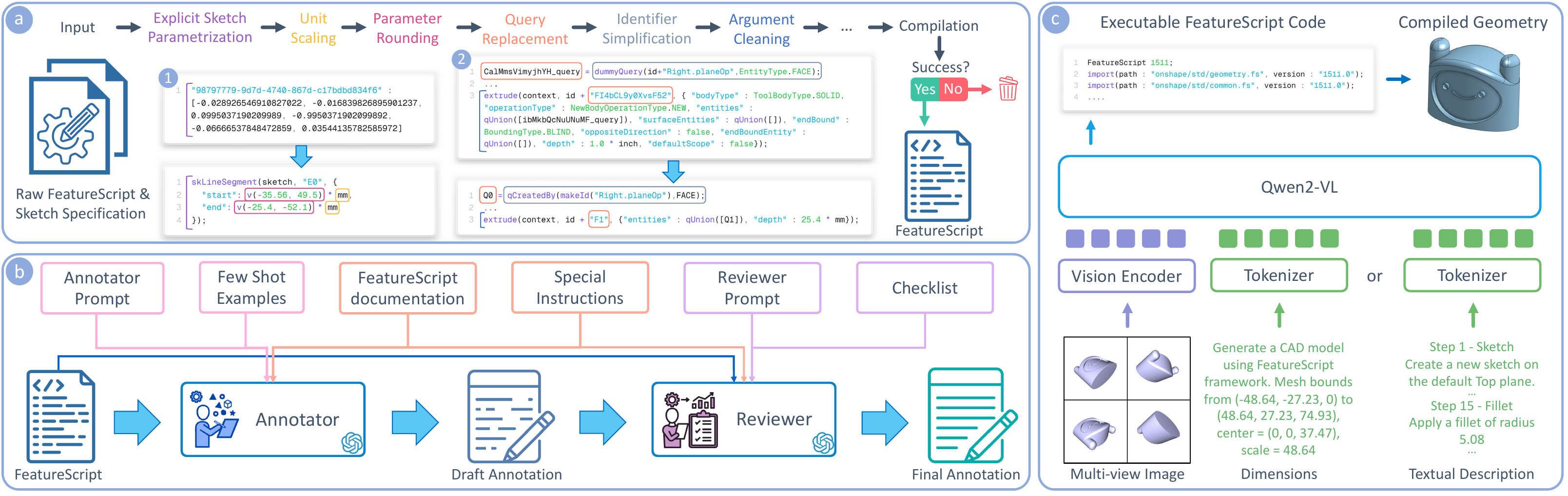}
    \caption{
        Method overview. (a) We propose a FeatureScript reconstruction pipeline to create a large dataset of CAD models with advanced modeling operations. (b) Each model is annotated with a textual description using our two-stage annotation pipeline. (c) We finetune Qwen2-VL conditioned on text and image inputs to generate FeatureScript code that can be directly compiled into a B-rep model.
    }
    \label{fig:method}
\end{figure*}

The foundation of our framework for generative CAD is a new representation of the design history.
Specifically, we propose to train LLM-based models to write code in FeatureScript language, used by the Onshape platform.
We develop a data acquisition pipeline that converts Onshape's internal CAD representation into clean, compact, and fully executable code suitable for training.
Using this pipeline, we collect a large set of CAD designs in the new representation and annotate each design with a textual description and multi-view images.
Finally, we adopt an existing VLM to generate CAD designs in the new representation and fine-tune it on the new dataset for text-conditioned generation and reconstruction from multi-view images.
\Cref{fig:method} shows an overview of our method.

\begin{figure*}[t]
    \centering
    \includegraphics[max height=14.5\baselineskip, max width=\textwidth]{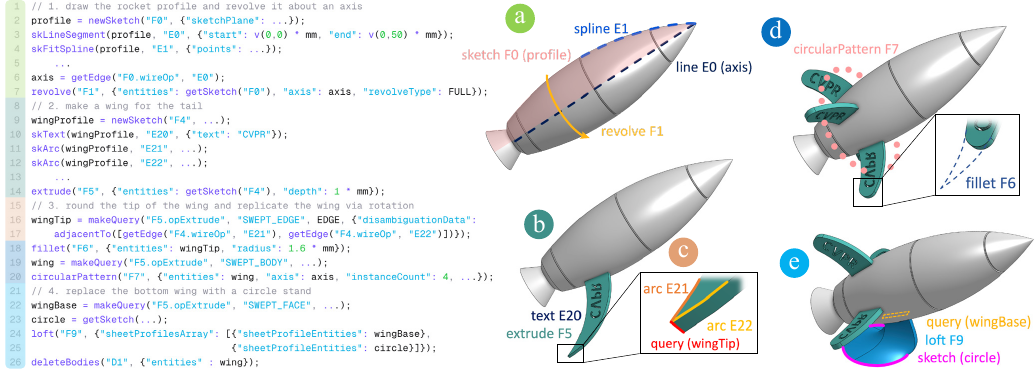}
    \caption{
        Example of FeatureScript code that describes the design history of the model shown on the right.
        (a) The profile of the rocket body is drawn using spline primitives and revolved about an axis.
        (b) The profile of the tail wing is drawn using arcs and text primitives and then extruded.
        (c) The wing tip is identified as the edge created by extruding the junction between the arcs of the wing profile.
        (d) The identified wing tip is rounded with a fillet, and the wing is replicated using a circular pattern.
        (e) A rectangular face of the wing is smoothly connected to a round base using a loft to form a stand for the rocket, and the corresponding wing is removed.
    }
    \label{fig:featurescript}
\end{figure*}

\subsection{FeatureScript code for generative CAD}
\label{sec:featurescript}

FeatureScript is the native language used within Onshape to define parametric modeling operations and 3D geometry.
It directly exposes the full set of modeling operations supported by the platform, including refinement and transformation operations missing in previous datasets.
In contrast to prior representations, FeatureScript enables robust and precise specification of operations such as fillet or loft via structured queries.
Furthermore, using FeatureScript as the training representation allows us to leverage a continuously growing corpus of real-world CAD designs authored by practicing engineers, without requiring lossy conversion steps.
As a result, our FeatureScript-based framework naturally reflects real industrial modeling practices, rather than synthetic workflows designed around representational limitations.

FeatureScript queries enable modeling operations to target existing geometric entities (edges, faces, bodies) defined by their origin, construction history, or relations to other entities.
\Cref{fig:featurescript} shows an example CAD model and its corresponding FeatureScript code.
In this example, queries, represented by the function \texttt{makeQuery} in lines 16, 19, and 22, are used
(1) to round the edge produced by a specific extrusion originating from the connection point between specific sketch primitives,
(2) to replicate the body produced by that extrusion using a circular pattern,
(3) to connect a specific face of this body to another face using a smooth transition,
and (4) to delete this body created only as an intermediate for subsequent operations.

The function \texttt{makeQuery} takes as input an operation identifier, a query type, an entity type, and disambiguation data.
The operation identifier scopes the query to a specific modeling operation within the feature history, \eg, the extrusion with id \texttt{F5}.
The query type encodes the topological role of the target entity within that operation, \eg, a side edge is \texttt{SWEPT\_EDGE}.
The entity type indicates the class of geometric entities targeted by the query, \ie, vertex, edge, face, or body.
Finally, the disambiguation data provides the additional information needed to resolve ambiguities and uniquely identify the intended entities when multiple candidates satisfy the query conditions.
The most common mechanisms are \textit{original set disambiguation} and \textit{topology disambiguation}, which specify either the entity's ancestors or its neighbors, as in \cref{fig:featurescript} line 17.
Together, these four parameters form a compact but expressive addressing scheme that enables robust references to individual geometric entities within a parametric CAD history.
Notably, this representation mirrors the way a human would verbally identify a geometric feature, \ie, by its origin, semantic role, categorical type, and distinguishing attributes.
This correspondence makes the queries particularly amenable to generation by large language models.
Overall, our representation preserves exact geometric and parametric fidelity while remaining compact and interpretable, making it well-suited for LLM-based generative modeling and downstream editing workflows.

\subsection{Data acquisition and the dataset}
\label{sec:dataset}

While FeatureScript is Onshape's native representation, for manually created models the platform provides only an internal representation that is neither directly executable nor interpretable.
This representation contains issues such as implicit parameterization, redundant or unused expressions, inconsistent unit usage, and randomly generated identifiers.
We develop a data acquisition pipeline that reconstructs high-quality, executable FeatureScript programs from Onshape's internal representation.

\Cref{fig:method.pipeline} illustrates the pipeline.
First, the sequence of modeling operations and their parameters are extracted.
Then, implicit and platform-dependent parameterization is replaced with explicit arguments.
For example, a line represented by a point and a direction vector is replaced with a start-end point definition, as illustrated with purple brackets in part (1) of \cref{fig:method.pipeline}.
The units of parameters are standardized to millimeters (\cref{fig:method.pipeline}, 1,~yellow).
Numeric expressions are resolved, and precision is standardized to two decimal places (\cref{fig:method.pipeline}, 1,~magenta).
Dummy queries are replaced with meaningful alternatives (\cref{fig:method.pipeline}, 2,~orange).
Random entity and variable names, which impede interpretation by LLMs, are replaced with compact, ordered identifiers (\cref{fig:method.pipeline}, 2,~grey).
Definitions of geometric operations are simplified and normalized (\cref{fig:method.pipeline}, 2,~blue).
Redundant operations and sketch entities that have no influence on the resulting geometry are removed.
Finally, each program is validated by checking that the code reproduces the original model.
Programs that fail validation are discarded (about 15\%).

We focus on 15 commonly used modeling operations, including sketch, extrude, revolve, sweep, loft, fillet, chamfer, shell, hole, boolean operations, patterns, \etc, discussed in \cref{sec:sup_ops}.
Unlike prior datasets restricted to sketch and extrude operations, this operation set covers the full modeling workflow, from initial geometry definition to refinement and reuse, mirroring real mechanical design practice.

Applying our pipeline to these 15 modeling operations, we collect a dataset of 451k CAD designs with design histories represented as concise, executable FeatureScript code.
Executing each program in Onshape reproduces the original real-world design, preserving full geometric and parametric fidelity.
This dataset forms the foundation for training LLM-based generative models capable of producing complex and editable CAD designs.

To maximize compatibility with prior work, we construct our dataset from source CAD designs that form a subset of the collection underlying the ABC dataset~\cite{koch2019abc} and a superset of the collection used by the DeepCAD~\cite{wu2021deepcad} and Text2CAD~\cite{wang2025texttocad} datasets.
This overlap enables direct comparison of generative methods across representations on shared geometry and supports gradual migration from earlier token-based or Python-based datasets to FeatureScript.

\subsection{Annotation with natural language descriptions}
\label{sec:annotation}

Generating CAD models from natural language descriptions enables intuitive, high-level specification of design intent.
However, accurate text-to-CAD generation requires clear, unambiguous descriptions of the construction steps.
Public CAD model repositories do not provide textual descriptions, while manual annotation is infeasible at scale.
Prior works~\cite{khan2024text2cad,xu2024cadmllm} therefore generate descriptions automatically from programmatic CAD representations using LLMs.
We adopt this strategy for our FeatureScript-based representation, which supports more accurate descriptions thanks to the structured and semantically rich nature of the code.

\Cref{fig:method.annotation} illustrates our annotation pipeline.
We build it around two LLMs, an \emph{Annotator} and a \emph{Reviewer}, whose roles are defined by system prompts (see in \cref{sec:sup_annotation.prompts}).
Given an input CAD model represented as FeatureScript code, the Annotator produces an initial draft description.
It receives the source code and few-shot examples and produces a structured summary of the construction process.
The Reviewer refines this draft and ensures consistency with the original code.
It checks the correctness of the operation sequence, resolves ambiguities, corrects terminology, and verifies numerical parameters.
Both models also receive FeatureScript documentation, along with interpretation guidelines and rules for phrasing and structuring the final output.

\begin{table*}[t]
    \centering
    \caption{
        Comparison of different frameworks in text-conditioned generation (top) and reconstruction from multi-view images (bottom).
        The models are trained with different representations of design history (Repr.), different text annotations (Annot.), and different CAD collections, where D denotes DeepCAD, R denotes CAD-Recode, and the number of modeling operations is shown in parentheses.
        The DeepCAD test set includes only sketch and extrude operations, while Our test set includes 15 different operations.
        Text2CAD annotations (T2C) are not available for the new designs, so baselines trained on these annotations cannot be evaluated for text-conditioned generation on the new test set.
        The \textbf{best} and \underline{second-best} results for each task are shown in bold and underlined, respectively.
    }
    \includegraphics[max height=50\baselineskip, max width=\textwidth]{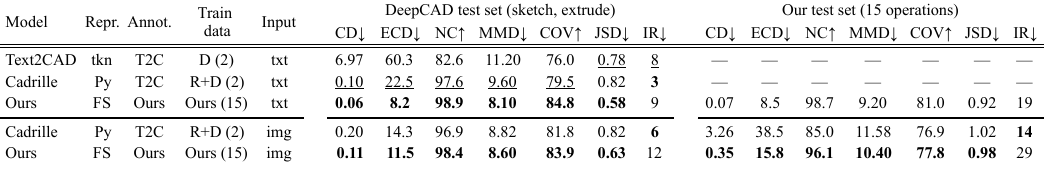}
    \label{tab:metrics_comp}
\end{table*}

Using our FeatureScript code representation as the basis for annotation significantly simplifies the process.
The code expresses modeling operations, parameters, and dependencies explicitly, making the construction sequence inherently interpretable.
This reduces the risk of missing or misrepresenting operations and helps maintain alignment between the textual description and the actual design logic.

Adding documentation into the prompt helps the models interpret operations that depend on geometric context.
Without it, LLMs easily misidentify reference entities (\eg, confusing interior and exterior contours or referencing incorrect edges or faces).
Providing documentation reduces such errors and leads to clearer and more precise descriptions.

The two-agent design improves accuracy and reliability.
The Annotator ensures global completeness and coherence, while the Reviewer focuses on correctness and detail.
This division reduces failure modes common in single-agent generation, such as operation omissions, incorrect naming, or inadvertent leakage of code into the text.
Few-shot examples improve consistency in style and structure across the dataset.

Together, these components yield accurate, interpretable, and well-structured textual descriptions.
Experiments show that training LLMs on our descriptions improves text-to-CAD generation performance, producing CAD models that match the design intent more precisely.
We additionally compare our descriptions to those from prior work in \cref{sec:sup_annotation.vs_t2c}.

\subsection{Learning FeatureScript code generation}
\label{sec:learning}

We aim to enhance existing vision-language models for CAD generation.
To this end, we adopt Qwen-VL~\cite{qwen2vl} to generate design histories in the new representation conditioned on text descriptions or multi-view images (\cref{fig:method.llm}).
Previous methods normalize CAD models to be zero-centered with a fixed scale for ease of generation.
In contrast, we train the model to generate designs in the natural dimensions specified by engineers.
To do so, we provide the model with the center of the design's bounding box and its extent through the text prompt for both text-conditioned and image-conditioned generation.
In \cref{sec:sup_results.scale_aware}, we show that providing these parameters improves generation accuracy in the new representation.

\section{Experiments}
\label{sec:experiments}

We evaluate our generative CAD framework by comparing a Qwen2-VL-2B model~\cite{qwen2vl} trained with our new representation against state-of-the-art approaches for text-conditioned generation and reconstruction from multi-view images.
We further perform ablations to assess the contributions of the proposed design-history representation, the extended set of modeling operations, and the improved textual annotations.

\subsection{Experimental setup}
\paragraph{Implementation details.}
We perform supervised fine-tuning of Qwen2-VL-2B in two stages.
First, we fine-tune the model on \raisebox{0.5ex}{\texttildelow}170k designs containing only sketch and extrude operations, corresponding to the DeepCAD dataset~\cite{wu2021deepcad}.
This stage establishes core geometric reasoning.
We then fine-tune the model on \raisebox{0.5ex}{\texttildelow}405k designs from our full dataset, enabling generalization to all 15 modeling operations.
In both stages, the model is conditioned either on textual descriptions or a \(2\times 2\)~grid of multi-view images, following Cadrille~\cite{kolodiazhnyi2025cadrille}.
We provide more details in \cref{sec:sup_implementation}.

\paragraph{Baselines.}
We compare against two leading frameworks: generation of tokenized sequences and generation of Python code, represented by Text2CAD and Cadrille.
Text2CAD~\cite{khan2024text2cad} generates tokenized CAD sequences of sketch and extrude operations conditioned on text.
It trains a 360M-parameter transformer using text annotations at four abstraction levels over the 170k DeepCAD designs.
Cadrille~\cite{kolodiazhnyi2025cadrille} generates CADQuery-based Python design histories from text, multi-view images, or point clouds.
It fine-tunes Qwen2-VL-2B via supervised fine-tuning (SFT) followed by reinforcement learning.
We compare our SFT model against the SFT version of Cadrille, trained on a mixture of DeepCAD designs and synthetic designs from CAD-Recode~\cite{rukhovich2024cadrecode}, for a total of \(1.17\)M training examples.
We additionally compare against a mesh generation method TRELLIS~\cite{xiang2025trellis} in \cref{sec:sup_comparison.trellis}.

\begin{figure}[t]
    \centering
    \includegraphics[max height=260\baselineskip, max width=\columnwidth]{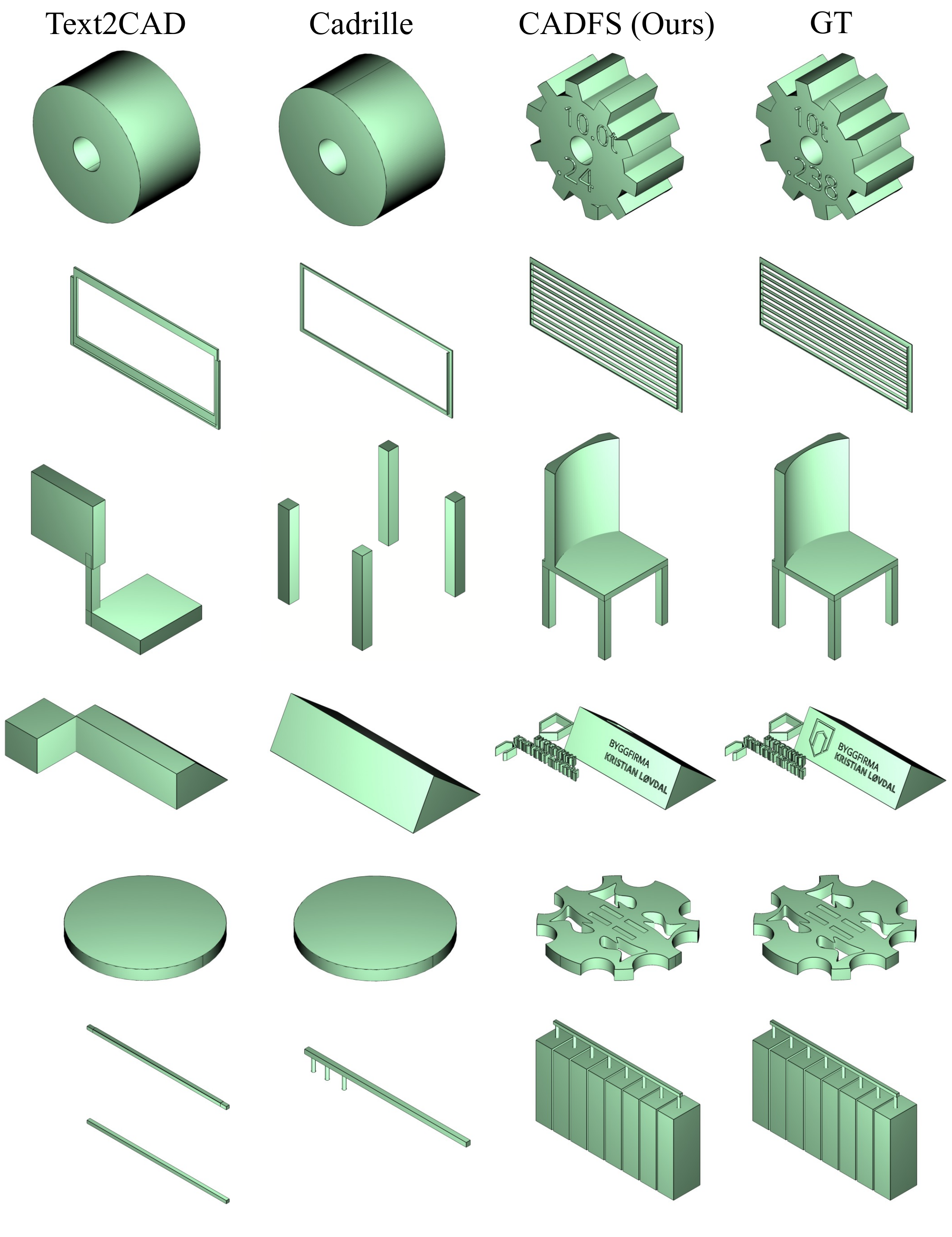}
    \caption{
        Qualitative comparison in text-conditioned CAD model generation on the DeepCAD test set.
    }
    \label{fig:txt_comp}
\end{figure}

\begin{figure}[t]
    \centering
    \includegraphics[max height=260\baselineskip, max width=\columnwidth]{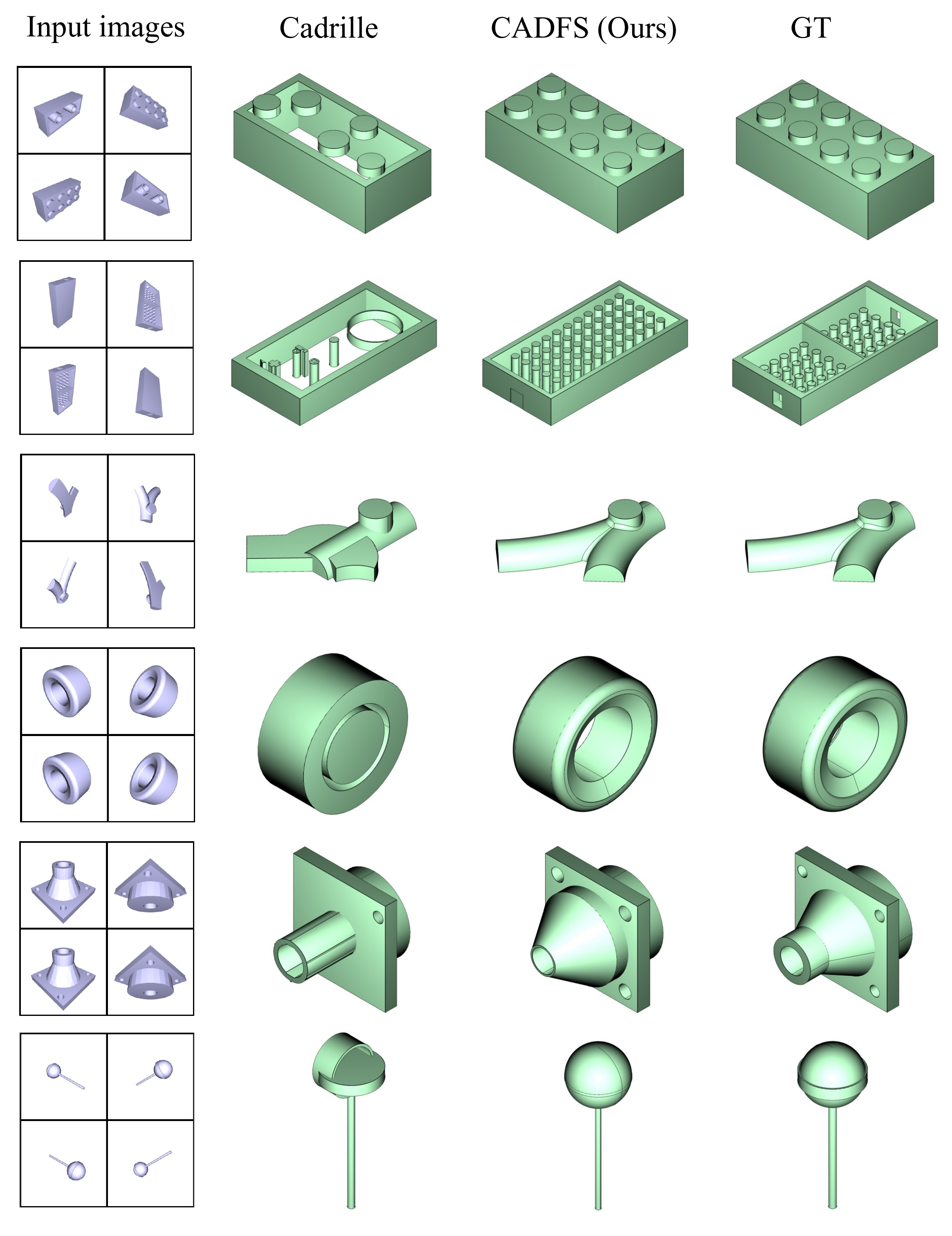}
    \caption{
        Qualitative comparison in multi-view CAD reconstruction.
        Rows 1--2 show results on the DeepCAD test set, and rows 3--6 show results on our test set.
    }
    \label{fig:img_comp}
\end{figure}

\paragraph{Data.}
We compare the models on two test sets.
To assess performance on designs constructed from sketch and extrude operations, we use the 7278 designs from the DeepCAD test set that have Text2CAD, CADQuery, and FeatureScript annotations available simultaneously.
To evaluate performance on the full set of modeling operations, we sample \raisebox{0.5ex}{\texttildelow}9k designs from our new dataset, covering all 15 operations.
The new test set is sampled such that, when combined with the DeepCAD test set, the distribution of operations matches that of the subset of ABC~\cite{koch2019abc} that contains designs expressible with our 15 operations.
We additionally compare the methods on the CADParser dataset~\cite{zhou2023cadparser} in \cref{sec:sup_comparison.cadparser}.

For text-conditioned generation, both baselines rely on Text2CAD textual annotations.
We evaluate each method with the annotations used in its training.
For multi-view reconstruction, we train and test all methods with a per-view resolution of \(256\times 256\).
We render the input images from the original Onshape geometry and also use this geometry as the reference for metrics.

\paragraph{Metrics.}
Following prior work on CAD design-history generation~\cite{wu2021deepcad,li2023secadnet,khan2024text2cad,rukhovich2024cadrecode,kolodiazhnyi2025cadrille}, we evaluate geometric accuracy and diversity.
For geometric accuracy, we compute Chamfer Distance (CD), Edge Chamfer Distance (ECD), and Normal Consistency (NC) between generated and reference CAD models.
To assess the fidelity of the geometric distribution, we use Minimal Matching Distance (MMD).
We measure the diversity using Coverage (COV) and Jensen-Shannon Divergence (JSD).
We also report the Invalidity Ratio (IR), which measures the fraction of generated designs that fail to construct.
We additionally report topology validity metrics in \cref{sec:sup_comparison.more_metrics}.
We provide full metric definitions in \cref{sec:sup_eval}.

\subsection{Comparison against other frameworks}
\label{sec:experiments_comp}

\Cref{tab:metrics_comp} shows the quantitative comparison of the frameworks, \cref{fig:txt_comp,fig:img_comp} show the qualitative comparisons in text-conditioned generation and multi-view reconstruction respectively.

Our model substantially outperforms the models based on prior frameworks on both text-conditioned generation and multi-view reconstruction.
On the DeepCAD test set (sketch and extrude only), it achieves markedly higher geometric accuracy and diversity than Text2CAD and Cadrille.
In text-conditioned generation, it improves CD and ECD by 40\% and 64\% over Cadrille, while also achieving better MMD, COV, and JSD.
The Invalidity Ratio is comparable to Text2CAD and slightly worse than Cadrille, likely due to the lack of FeatureScript code in the VLM pretraining data compared to Python.

\begin{table*}[t]
    \centering
    \caption{
        Ablation study of our framework in text-conditioned generation (top) and reconstruction from multi-view images (bottom).
        We evaluate the individual contributions of the new representation (Repr.), the new text annotations (Annot.), and training on the new modeling operations (D denotes DeepCAD, R denotes CAD-Recode, and the number of modeling operations is shown in parentheses).
        The DeepCAD test set includes only sketch and extrude operations, while Our test set includes 15 different operations.
        Text2CAD annotations (T2C) are not available for the new designs, so models trained on these annotations cannot be evaluated for text-conditioned generation on the new test set.
        The \textbf{best} and \underline{second-best} results for each task are shown in bold and underlined, respectively.
    }
    \includegraphics[max height=50\baselineskip, max width=\textwidth]{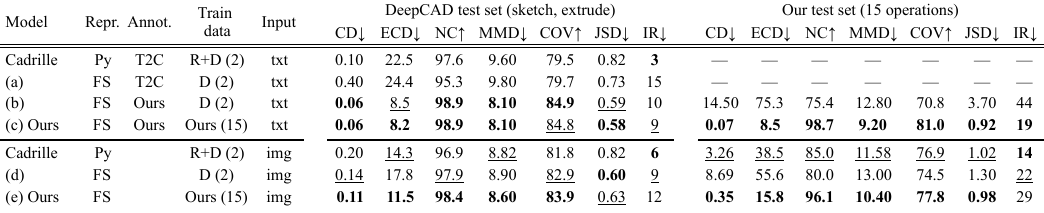}
    \label{tab:metrics_abl}
\end{table*}

On the new test set with the extended set of modeling operations, our model surpasses Cadrille in multi-view reconstruction in both accuracy and diversity.
Both models perform worse than in text-conditioned generation, highlighting the difficulty of design history reconstruction from images alone.
The baselines cannot be evaluated for text-conditioned generation on the new test set, as Text2CAD annotations are unavailable for the new designs.
Comparing our model across the two test sets shows that it generalizes well: the geometric accuracy remains consistent, while the diversity and validity exhibit a small drop attributable to the higher complexity and variability of the designs with the extended operation set.
Overall, our framework enables the generation of substantially more complex and dependency-rich CAD designs than prior frameworks.

\subsection{Ablation study}
\label{sec:experiments_ablation}

\Cref{tab:metrics_abl} shows the evaluation of contributions of individual components of our framework: the FeatureScript representation, the extended modeling operation set, and the representation-aligned textual annotations.

Comparing Python-based Cadrille with the FeatureScript-based models (a) and (d) isolates the effect of the representation (note that all three models are based on Qwen2-VL-2B).
Using FeatureScript with the DeepCAD training set and Text2CAD annotations yields performance on par with Cadrille despite its significantly larger training corpus.
This shows that FeatureScript is a viable alternative to the Python-based representation.
At the same time, using FeatureScript allows us to scale training to complex real-world designs with a broad range of modeling operations, which yields a substantial performance improvement, as shown by comparing models (a)~and~(c) or (d)~and~(e).

Comparing models (a)~and~(b) isolates the effect of the new textual annotations.
The new annotations provide clear gains in geometric accuracy and diversity, highlighting the importance of annotations aligned with the underlying design-history representation.
We provide additional ablation results in \cref{sec:sup_results}.

\section{Conclusions}
\label{sec:conclusions}
We introduced CADFS, a data-centric framework that enables large vision-language models to generate complex CAD design histories spanning a broad range of modeling operations.
By leveraging a new FeatureScript-based representation, a large-scale dataset of 450k real-world designs, and representation-aligned textual annotations, our approach overcomes the limitations of prior tokenized and Python-based CAD workflows.

A VLM trained within this framework achieves state-of-the-art performance in both text-conditioned generation and multi-view reconstruction, producing more accurate, diverse, and feature-rich CAD models than prior methods.
Our ablations confirm that each component of the framework --- the representation, the extended operation set, and the textual annotations --- contributes substantially to these gains.

Our FeatureScript reconstruction pipeline can be used directly to scale the dataset to newly added designs in the Onshape library.
Overall, CADFS significantly broadens the complexity and fidelity achievable in generative CAD, opening the door to models capable of producing realistic, editable designs aligned with real engineering practice.

\paragraph{Acknowledgments.}
We are grateful to Onshape for providing public access to a vast library of CAD designs.
The presented results were obtained with the use of the supercomputer Zhores~\cite{zacharov2019zhores}.
The work was supported by the grant for research centers in the field of AI provided by the Ministry of Economic Development of the Russian Federation in accordance with the agreement 000000C313925P4F0002 and the agreement №139-10-2025-033.

    \clearpage
\maketitlesupplementary

In \cref{sec:sup_results}, we present additional ablation studies on key design choices in our framework, including
(1) the choice of the base VLM used for fine-tuning,
(2) the contribution of individual components of our FeatureScript-based representation,
(3) the effect of combining the design history representation with appropriate textual annotations,
(4) the effect of providing the model with explicit design dimensions during generation,
and (5) the impact of input image resolution on reconstruction quality for models involving the newly introduced operations.
In \cref{sec:sup_comparison}, we report evaluation with additional metrics, on additional data, and an additional comparison with a mesh-based approach.
In \cref{sec:sup_failure}, we discuss failure cases of our model.
In \cref{sec:sup_implementation}, we provide implementation details.
In \cref{sec:sup_ops}, we discuss our choice of modeling operations.
In \cref{sec:sup_annotation}, we provide further details of our annotation procedure.
In \cref{sec:sup_eval}, we describe the evaluation details.

\section{Additional ablation studies}
\label{sec:sup_results}

\subsection{Choice of base model}

In our main experiments, we use the Qwen2-VL-2B model~\cite{qwen2vl}.
Here, we compare it with the larger Qwen3-8B variant~\cite{yang2025qwen3} on text-conditioned generation using the DeepCAD test subset.
As shown in \cref{tab:metrics_8b}, the 2B model performs on par with the 8B model, while requiring roughly half the training time.
These results indicate that Qwen2-VL-2B achieves a strong balance between computational efficiency and generation accuracy.

\subsection{FeatureScript representation}

The Onshape platform does not directly provide clean, executable FeatureScript code.
We reconstruct high-quality FeatureScript programs from Onshape's internal representation using our data acquisition pipeline.
\Cref{tab:metrics_representation} reports an ablation over progressively refined variants of this representation.
In these experiments, we train and test Qwen3‑8B for text‑conditioned generation on the DeepCAD subsets.

We begin with minimally processed executable code extracted from the internal Onshape representation paired with abstract Text2CAD~\cite{khan2024text2cad} annotations (a).
This baseline yields low generation accuracy, indicating that the raw Onshape representation is insufficient for precise generation.

As the first refinement, we replace the arbitrary identifiers of geometric entities (edges, faces, bodies) with compact deterministic ones (\eg, \textquote{F0}, \textquote{E0}, \textquote{E1} in \cref{fig:featurescript}) (b).
This improves geometric accuracy (Chamfer Distance) and diversity (COV) by $22\%$ and $13\%$, respectively.

Next, we train a model on expert Text2CAD annotations (c), which substantially improves performance across all metrics.
However, this also doubles the Invalidity Ratio.
This suggests that the detailed expert descriptions from Text2CAD are misaligned with the entangled internal Onshape representation of CAD models.

To address this, we replace implicit definitions of sketch elements with explicit ones (d), \eg, representing line segments by their endpoints rather than by an origin point and direction.
This significantly reduces the Invalidity Ratio from $24\%$ to $10\%$ while maintaining accuracy and diversity.

We further disentangle the representation by simplifying modeling operations (e), yielding an additional $21\%$ reduction in Chamfer Distance.

With the FeatureScript code now concise and interpretable, we generate our own textual annotations tailored to this representation and describing the CAD models more precisely (f).
This key alignment between representation and annotations produces a $60\%$ reduction in Chamfer Distance and improvements across other metrics as well.

Finally, we standardize numerical precision to two decimal places (g).
While this has only a minor impact on performance, it reduces code length and improves consistency.

\begin{table}[t]
    \centering
    \caption{
        Comparison of LLMs of different size trained for text-conditioned CAD generation based on our framework.
        The \textbf{best} result is shown in bold text.
    }
    \includegraphics[max height=50\baselineskip, max width=\columnwidth]{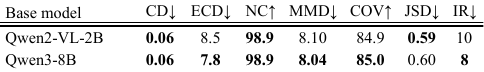}
    \label{tab:metrics_8b}
\end{table}

\subsection{Different combinations of representations and annotations}

\Cref{tab:metrics_more_abl} shows an additional evaluation of the contributions of our FeatureScript representation and our new textual annotations.
We compare models trained with our standardized FeatureScript representation paired with different annotations: (a) T2C short annotations, (b) T2C expert annotations, and (e) ours.
We also compare models trained with our annotations paired with different design history representations: (c) Python code, (d) the minimally processed \textquote{raw} FeatureScript, and (e) the standardized FeatureScript.
In these experiments, we train and test Qwen2-VL-2B on the DeepCAD subsets.

\begin{table*}[t]
    \centering
    \caption{
        Ablation study of our FeatureScript-based representation.
        We compare text-conditioned models trained using progressively refined variants of this representation.
        The \textbf{best} and \underline{second best} results are shown in bold text and underlined respectively.
    }
    \includegraphics[max height=50\baselineskip, max width=\textwidth]{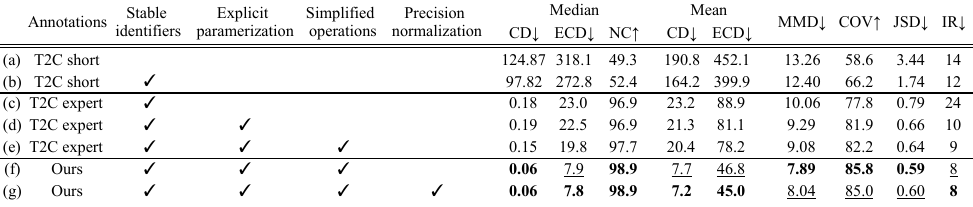}
    \label{tab:metrics_representation}
\end{table*}

\begin{table}[t]
    \centering
    \caption{Comparison of different combinations of representations and annotations.}
    \includegraphics[max height=50\baselineskip, max width=\columnwidth]{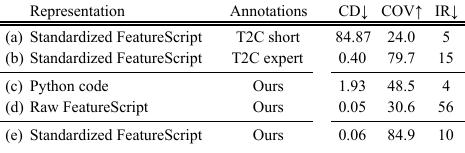}
    \label{tab:metrics_more_abl}
\end{table}

Only pairing the standardized FeatureScript with our annotations (e) simultaneously achieves low geometric error (CD), high diversity (COV), and a low invalid generation rate (IR).
Pairing FeatureScript with the Text2CAD annotations (a,b), or our annotations with Python (c) leads to significantly higher geometric error.
This is because these annotations describe the design process in substantially different representations.
The Text2CAD annotations and Python scripts are derived from the DeepCAD command sequence, while our annotations are derived from FeatureScript code.
For example, in DeepCAD, sketches are created in a normalized coordinate frame and then scaled and translated, whereas in FeatureScript everything is modeled directly in a global coordinate frame.

Training on raw FeatureScript, even with our annotations (d), results in a high fraction of invalid outputs (IR) due to the high degree of entanglement in the raw FeatureScript.

\subsection{Scale-aware multi-view reconstruction of CAD}
\label{sec:sup_results.scale_aware}

Generating CAD design histories as code naturally enables models to operate directly in the physical units used by engineers.
Prior code-based frameworks, however, inherit normalized coordinate systems from token-sequence representations, preventing generation at real-world scales.
In contrast, our CAD programs preserve each design’s original dimensions, allowing models to learn and generate geometry at true scale.
Our text annotations follow the same principle and specify all measurements in native units.

\begin{table}[t]
    \centering
    \caption{
        Comparison of multi-view CAD reconstruction models trained with and without additional information about the bounding box dimensions of the design.
        The \textbf{best} result is shown in bold text.
    }
    \includegraphics[max height=50\baselineskip, max width=\columnwidth]{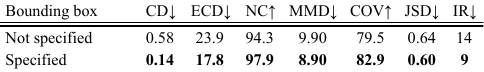}
    \label{tab:metrics_bbox}
\end{table}

In this context, reconstruction from multi-view images presents an additional challenge: absolute scale cannot be reliably inferred from visual input alone.
To mitigate this, we provide the VLM with the bounding-box dimensions and position of each design as part of the textual prompt
(\eg, \textit{\textquote{Generate a CAD model using FeatureScript framework. Bounds from (-114.66, -69.35, -31.78) to (68.33, 76.26, 50.8), center = (-23.17, 3.45, 9.51), scale = 91.5}}).

\Cref{tab:metrics_bbox} compares models trained with and without this scale information on the DeepCAD test set.
Supplying the bounding-box parameters significantly improves geometric accuracy: for example, improves Chamfer Distance by $76\%$ and reduces the Invalidity Ratio by $36\%$.

\subsection{Image resolution}

We additionally examine the effect of input image resolution on CAD reconstruction quality.
\Cref{tab:metrics_resolution} compares a model trained with a per-view resolution of \(256\times 256\) against one trained with \(128\times 128\).
We evaluate both on the DeepCAD test set and on our new test set containing the full set of modeling operations.

On the DeepCAD test set, which consists primarily of simple sketch-and-extrude geometry, the gains from higher resolution are modest.
In contrast, on our test set featuring more complex geometric structures, the benefits are substantially larger.
Overall, doubling the input resolution yields roughly a \(2\times\) improvement in Chamfer Distance and a $23\%$ improvement in Edge Chamfer Distance.
These results further highlight that models trained on richer, multi-operation CAD data are better positioned to leverage higher-fidelity visual input, reflecting the greater geometric diversity and complexity of our new data.

\begin{table*}[t]
    \centering
    \caption{
        Comparison of multi-view CAD reconstruction models trained with different per-view image resoluton.
        The \textbf{best} result is shown in bold text.
    }
    \includegraphics[max height=50\baselineskip, max width=\textwidth]{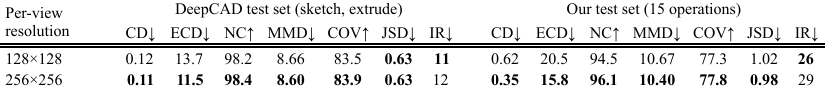}
    \label{tab:metrics_resolution}
\end{table*}

\begin{table*}[t]
    \centering
    \caption{
        Additional comparison of our framework with a Python code-based Cadrille in text-conditioned generation (left) and multi-view reconstruction (center and right) on topology validity metrics.
        The \textbf{best} result is shown in bold text.
    }
    \includegraphics[max height=50\baselineskip, max width=\textwidth]{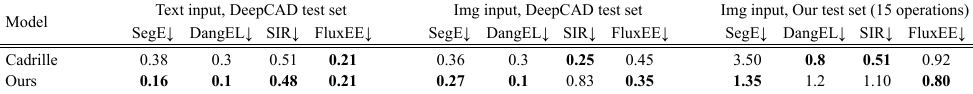}
    \label{tab:metrics_topo}
\end{table*}

\section{Additional comparisons}
\label{sec:sup_comparison}

\subsection{Additional metrics}
\label{sec:sup_comparison.more_metrics}

\Cref{tab:metrics_topo} reports additional topology validity metrics from~\cite[Sec.~6.1.4]{xu2024cadmllm}: Segment Error (SegE), Dangling Edge Length (DangEL), Self-Intersection Ratio (SIR), and Flux Enclosure Error (FluxEE).
Our model achieves high topological validity.

\subsection{Additional test data}
\label{sec:sup_comparison.cadparser}

In \cref{tab:metrics_cadparser}, we compare our model with Cadrille in image-based reconstruction on the CADParser dataset~\cite{zhou2023cadparser}, which features 5 operations: sketch, extrude, revolve, fillet, and chamfer.
The dataset includes ~40k designs obtained from an initial set of ~6.8k base designs via augmentation.
We test the models on a subset of 6.8k designs originating from different base designs.

The results are consistent with those on the DeepCAD and CADFS test sets.
Our model achieves significantly higher accuracy (CD, ECD) than Cadrille, with comparable diversity (COV, JSD).

\subsection{Comparison with mesh-based generation}
\label{sec:sup_comparison.trellis}

To show that CAD model generation requires specialized methods, we compare our framework against the polygonal mesh generation method TRELLIS~\cite{xiang2025trellis} on the image-conditioned generation task.
In this comparison, TRELLIS takes an isometric CAD image as input and generates a mesh, followed by postprocessing.

\begin{figure}[t]
    \centering
    \includegraphics[max height=260\baselineskip, max width=\columnwidth]{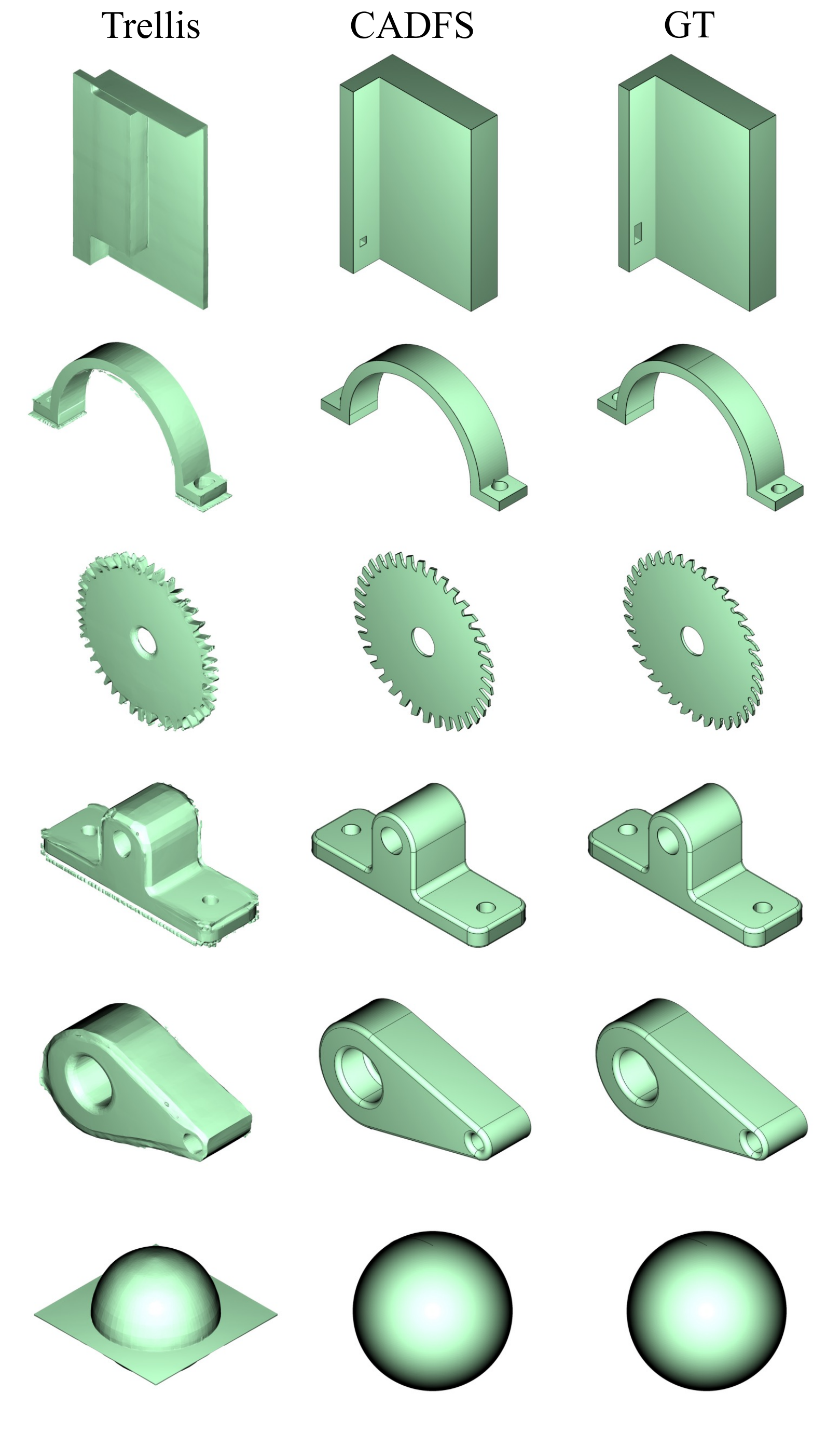}
    \caption{
        Qualitative comparison of our CAD-specific method and a mesh-based method of reconstruction from multi-view images.
    }
    \label{fig:mesh_comparison}
\end{figure}

\Cref{tab:metrics_mesh} and \cref{fig:mesh_comparison} show the quantitative and qualitative results, respectively.
The quantitative results show that our framework outperforms TRELLIS by a large margin.
The visual results show that our CAD-specific approach generates precise geometry, while the mesh-based method produces non-watertight meshes, over-smoothed edges, disconnected geometric components, and noisy surfaces.

\begin{table}[t]
    \centering
    \caption{
        Additional comparison of our framework with a Python code-based Cadrille in multi-view reconstruction on the CADParser dataset (5 operations).
        The \textbf{best} result is shown in bold text.
    }
    \includegraphics[max height=50\baselineskip, max width=\columnwidth]{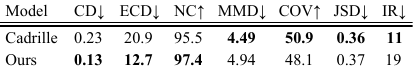}
    \label{tab:metrics_cadparser}
\end{table}

\begin{table*}[t]
    \centering
    \caption{
        Comparison of our CAD-specific method and a mesh-based method of reconstruction from multi-view images.
        The \textbf{best} result is shown in bold text.
    }
    \includegraphics[max height=50\baselineskip, max width=\textwidth]{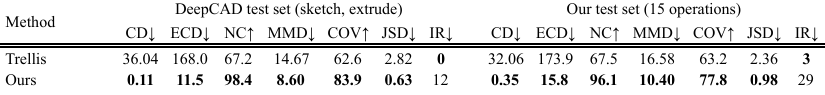}
    \label{tab:metrics_mesh}
\end{table*}

\begin{figure}[t]
    \centering
    \includegraphics[max height=260\baselineskip, max width=\columnwidth]{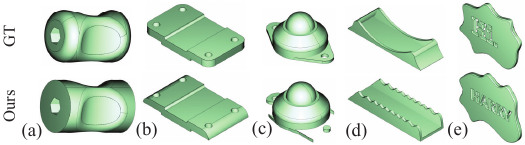}
    \caption{Examples of failure cases for the model trained on our data for text (a-c) and image input (d,e).}
    \label{fig:failure}
\end{figure}

\section{Failure cases}
\label{sec:sup_failure}

\Cref{fig:failure} shows examples of failure cases for the model trained on our data.
In both text- and image-conditioned modes, the model occasionally omits refinement operations such as fillets (a) or applies them to incorrect geometric entities (b).
It also occasionally produces inaccurate reconstructions overall (c,d).
In image-conditioned mode, the model often produces incorrect text (e), which we attribute to limitations of the Qwen-VL visual encoder.
It also frequently resorts to lower-level operations, \eg, multiple sketch-extrudes instead of a pattern.
This is likely due to the prevalence of simple operations in Onshape designs.

\section{Implementation details}
\label{sec:sup_implementation}

We train our model in two stages.
First, we fine-tune the model on \raisebox{0.5ex}{\texttildelow}170k designs containing only sketch and extrude operations, corresponding to the DeepCAD dataset~\cite{wu2021deepcad}.
This stage establishes core geometric reasoning.
We then fine-tune the model on \raisebox{0.5ex}{\texttildelow}405k designs from our full dataset that remain after excluding the test splits and scripts longer than 8192 tokens.
This enables generalization to all 15 modeling operations.
In both stages, the model is conditioned either on textual descriptions or a \(2\times 2\)~grid of multi-view images, following Cadrille~\cite{kolodiazhnyi2025cadrille}.
In the first stage, we use images at a resolution of \(128 \times 128\).
In the second stage, we increase the image resolution to $256 \times 256$ to improve geometric reasoning on complex structures, while keeping all other hyperparameters unchanged.
At each stage, we train the model for 3 epochs with a batch size of 128, using the Adam-W optimizer~\cite{loshchilov2017decoupled} with an initial learning rate of~$2\text{e-4}$, a linear warmup ratio of~\(0.05\), and a cosine decay schedule.
Training on 8 NVIDIA A100 GPUs with DeepSpeed~\cite{rasley2020deepspeed}, FlashAttention-2~\cite{dao2023flashattention2}, and Liger-Kernel~\cite{hsu2025ligerkernel} optimizations takes 30 and 76 hours for the first and second stages, respectively, using 24~GB of VRAM per GPU.

\section{Choice of modeling operations}
\label{sec:sup_ops}

\paragraph{Sketch-based construction of primary solids.}
\emph{Sketch} defines 2D profiles composed of lines, circles, arcs, ellipses, elliptical arcs, Bezier curves, splines, and text.
This expands beyond prior datasets limited to lines, circles, and circular arcs.
Sketches serve as the foundation for most solids.
\emph{Extrude} creates 3D solids by extending sketch profiles linearly, commonly used for prismatic parts and structural components.
Unlike prior datasets, the FeatureScript representation enables extruding separate parts of a sketch.
\emph{Revolve} sweeps a sketch profile around an axis to create rotationally symmetric solids (\eg, shafts, housings, knobs).
\emph{Sweep} moves a profile along a spatial curve to form tubing, wire guides, and ergonomic handles.
\emph{Loft} interpolates smoothly between multiple profiles to create aerodynamic or freeform transitions.
\emph{Construction plane} defines reference planes used to position sketches, splits, and mirror operations.
These operations create the core massing geometry of a part.

\paragraph{Refinement and edge treatment.}
\emph{Fillet} rounds sharp edges to reduce stress concentrations, improve manufacturability, and meet ergonomic requirements.
\emph{Chamfer} replaces edges with straight bevels for deburring, clearance, or assembly guidance.
Both require selecting specific edges or faces from the evolving model.
Their inclusion is enabled by FeatureScript's geometric query mechanism.

\paragraph{Solid modification and material removal.}
\emph{Shell} hollows a solid part while maintaining structural walls.
\emph{Hole} creates parametric holes with standardized diameters, countersinks, and threads.
\emph{Boolean union, subtract, intersect} combine or remove solids to form complex assemblies or cutouts.
\emph{Delete body} removes construction intermediates or temporary helper geometry.
These operations support both constructive and subtractive manipulation of solids.

\paragraph{Replication and spatial reuse.}
\emph{Circular pattern} repeats features radially around an axis (\eg, bolt circles, gear spokes).
\emph{Mirror} produces symmetric geometry efficiently by reflecting features across planes.
\emph{Transform} applies rigid translations and rotations to reposition or duplicate bodies or features.
These operations capture the hierarchical, parametric reuse patterns common in engineered components.

\section{Annotation details}
\label{sec:sup_annotation}

\subsection{System prompts}
\label{sec:sup_annotation.prompts}

\begin{figure*}[p]
    \centering
    \includegraphics[max width=\textwidth]{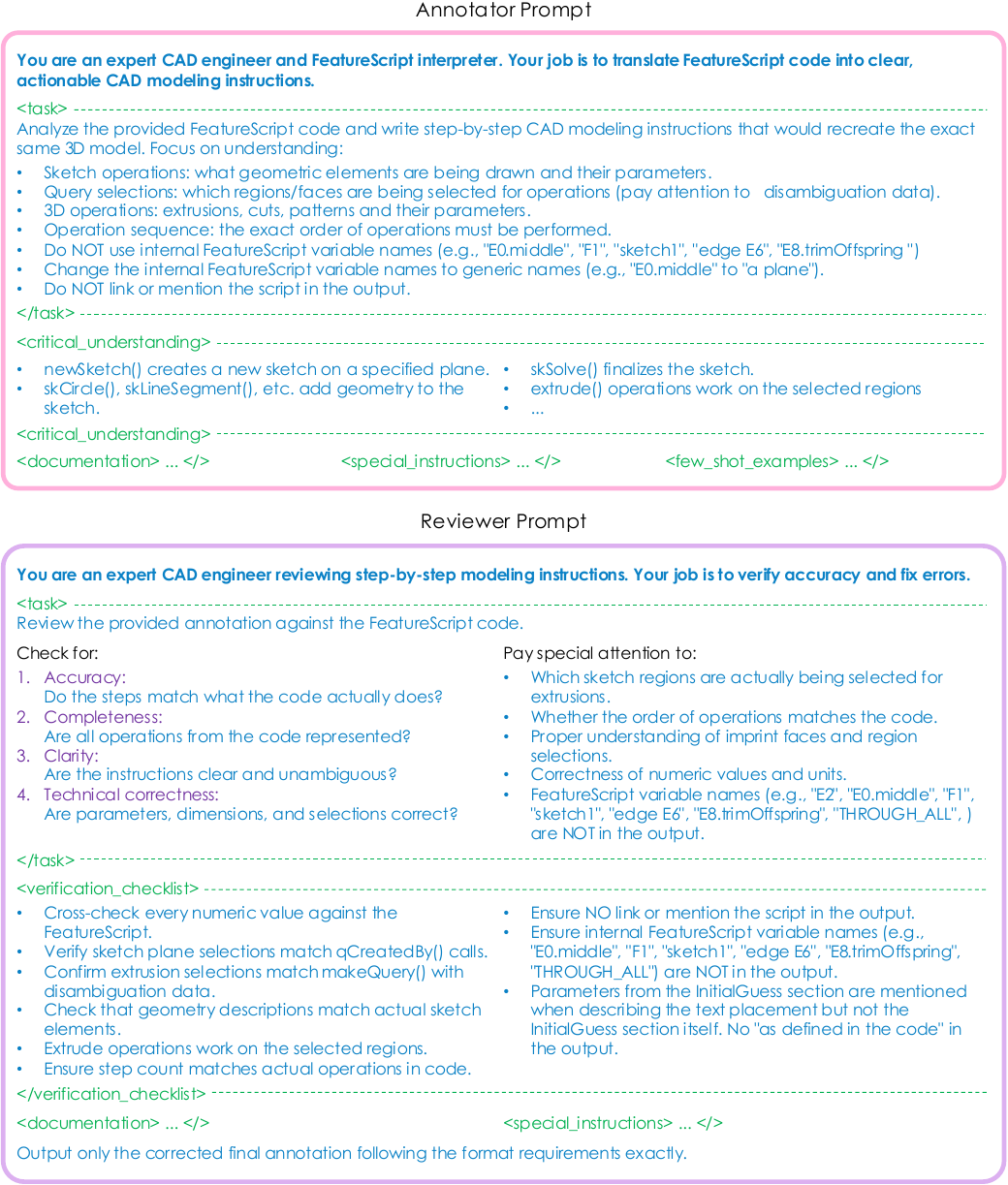}
    \caption{System prompts for the Annotator and Reviewer LLMs.}
    \label{fig:prompts_1}
\end{figure*}

\begin{figure*}[p]
    \centering
    \includegraphics[max width=\textwidth]{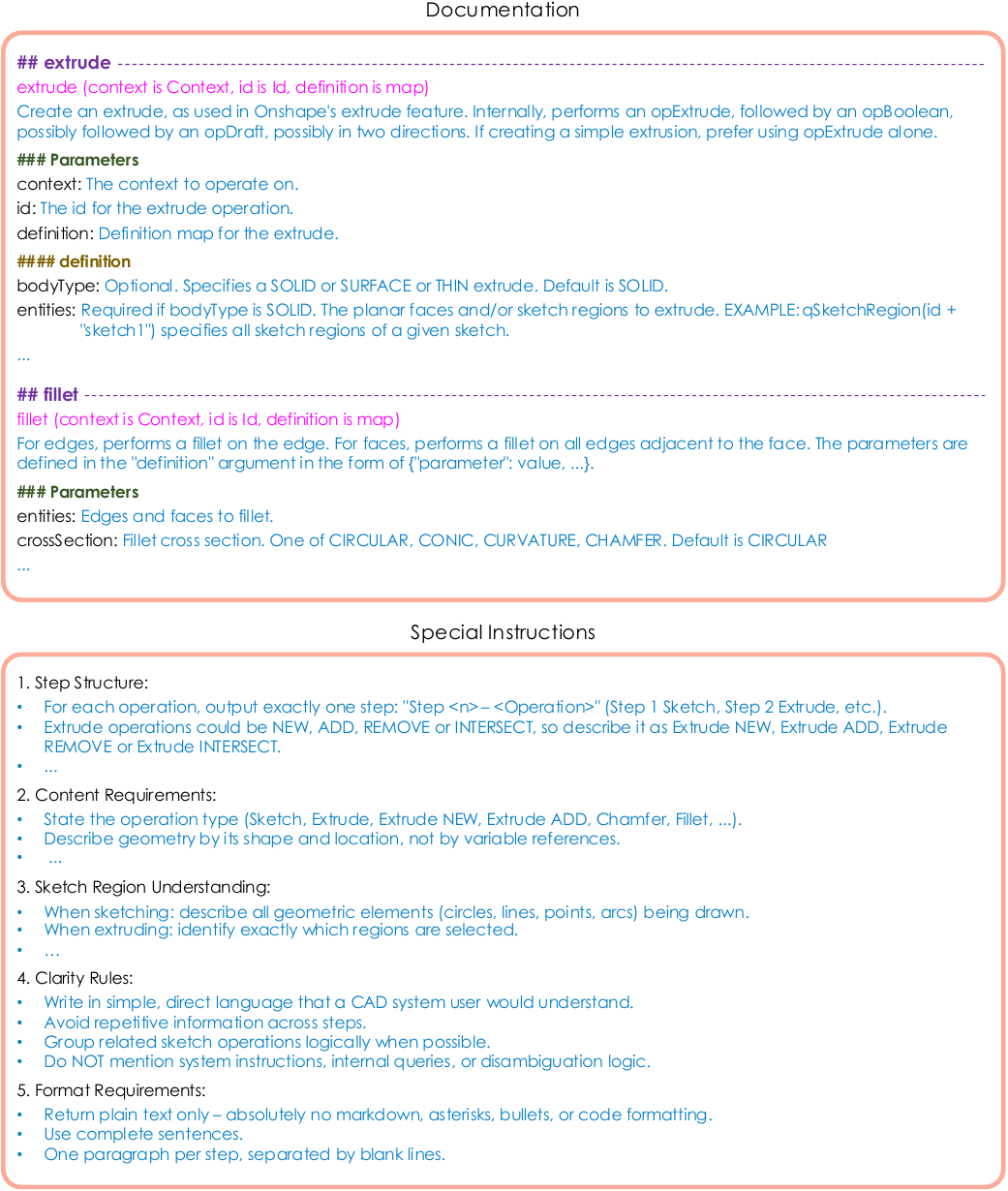}
    \caption{Excerpts from the FeatureScript documentation and special instructions provided to the Annotator and Reviewer LLMs.}
    \label{fig:prompts_2}
\end{figure*}

\Cref{fig:prompts_1} shows the system prompts for the Annotator and Reviewer LLMs.
The Annotator is instructed to translate the code representation of a CAD model into a natural language description.
The \texttt{critical\_understanding} section highlights the key characteristics of our FeatureScript representation, enabling the Annotator to develop a comprehensive understanding of the geometry expressed in the code and its construction process.
The Reviewer is instructed to perform a thorough validation of the Annotator's output.
Both models also receive FeatureScript documentation and special instructions for phrasing and structuring the final output, illustrated in \cref{fig:prompts_2}.

\subsection{Implementation details}

For both the Annotator and Reviewer LLMs, we use the gpt-oss-120B model~\cite{openai2025gptoss120b} with the Medium thinking configuration, which provides a good trade-off between annotation quality and computational requirements.
The annotation process for 450k CAD designs takes 7.5 days on 2 NVIDIA H100 GPUs.

\subsection{Comparison with Text2CAD annotations}
\label{sec:sup_annotation.vs_t2c}

Similar to our work, Text2CAD automatically generates textual descriptions from CAD representations using LLMs.
However, Text2CAD operates on simplified tokenized sequences derived from the original CAD data.
This conversion inevitably discards structural and geometric information, leading to incomplete or inaccurate prompts.
In contrast, we generate descriptions directly from the native CAD representation, enabling more faithful, detailed, and semantically aligned annotations.

\begin{figure*}[p]
    \centering
    \includegraphics[max width=\textwidth]{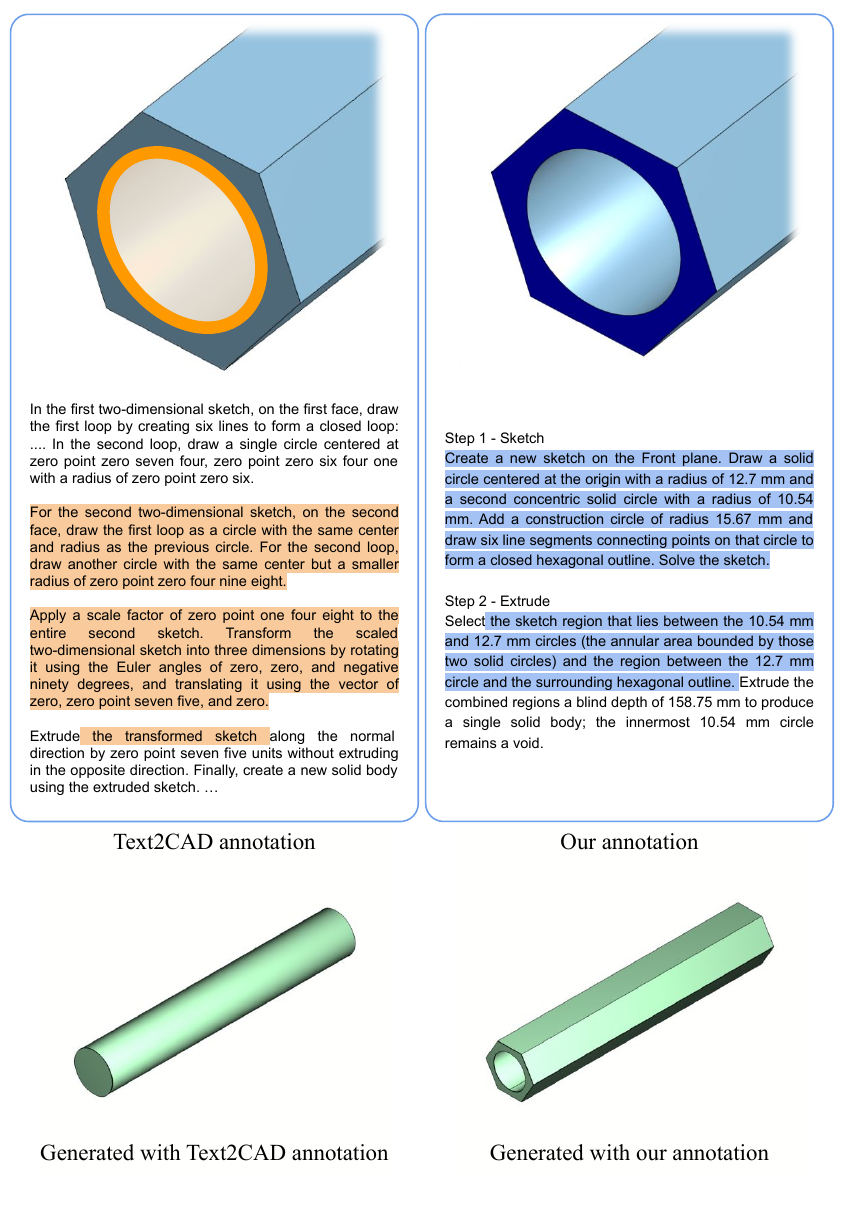}
    \caption{
        Comparison of the expert Text2CAD textual annotation (left) and our annotation (right) for the CAD model shown at the top.
        The CAD designs generated by models trained on each type of annotation are shown at the bottom.
        The Text2CAD annotation only describes extrusion of the second sketch with the circles (highlighted in orange) and forgets the first sketch with the hexagon.
        Our annotation describes the extrusion region correctly (highlighted in blue), supporting more accurate and complete CAD model generation.
    }
    \label{fig:annot_comparison_1}
\end{figure*}

\begin{figure*}[p]
    \centering
    \includegraphics[max width=\textwidth]{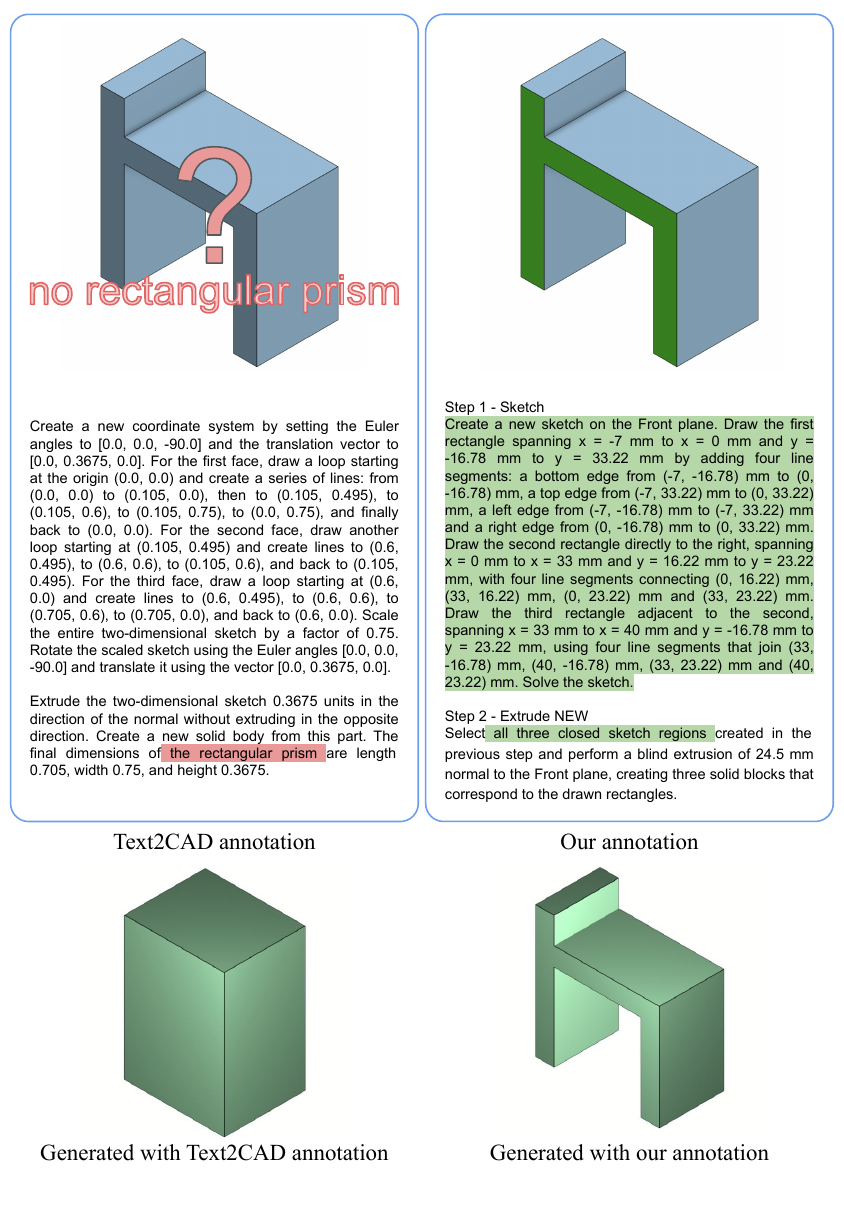}
    \caption{
        Comparison of the expert Text2CAD textual annotation (left) and our annotation (right) for the CAD model shown at the top.
        The CAD designs generated by models trained on each type of annotation are shown at the bottom.
        The Text2CAD annotation incorrectly describes the CAD model as a rectangular prism, which leads to the corresponding inaccurate generation result.
        Our annotation is consistent with the model geometry and leads to a correct result.
    }
    \label{fig:annot_comparison_2}
\end{figure*}

\begin{figure*}[p]
    \centering
    \includegraphics[max width=\textwidth]{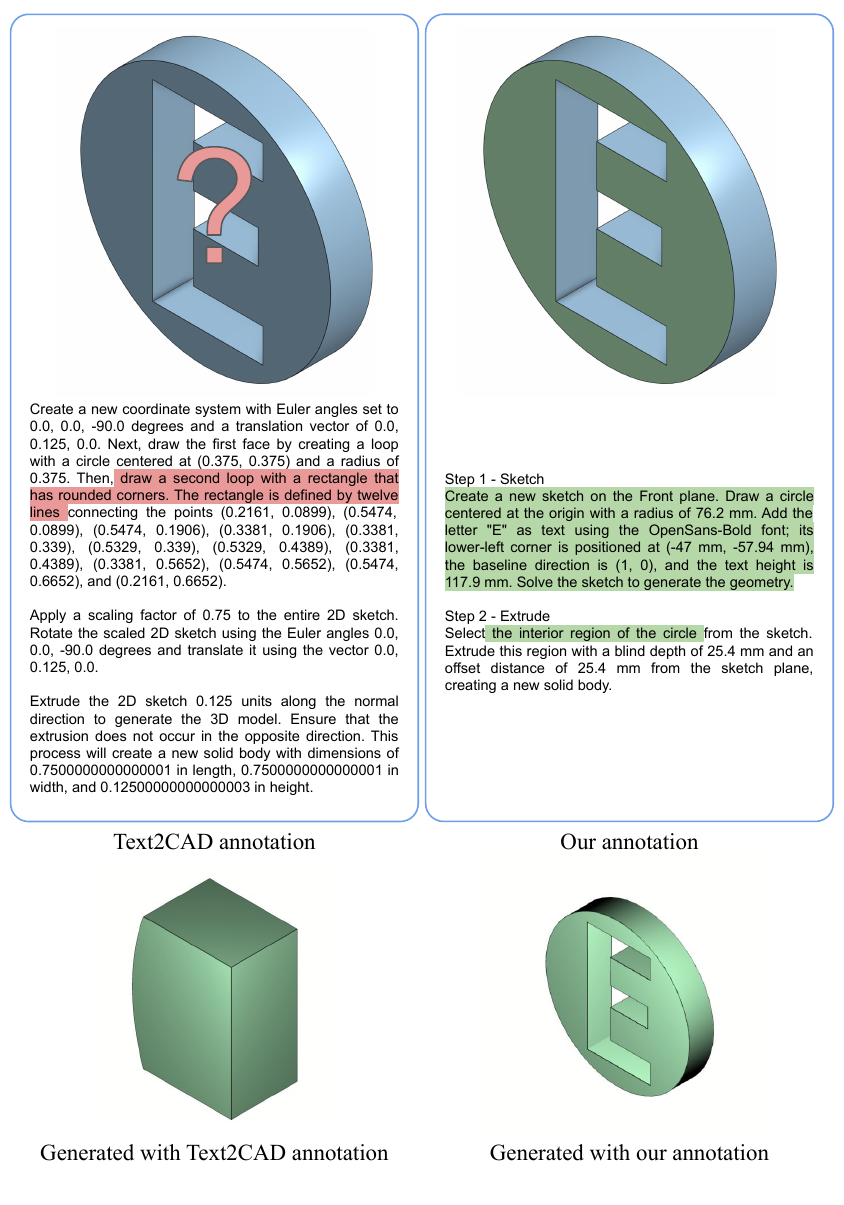}
    \caption{
        Comparison of the expert Text2CAD textual annotation (left) and our annotation (right) for the CAD model shown at the top.
        The CAD designs generated by models trained on each type of annotation are shown at the bottom.
        The Text2CAD annotation incorrectly describes the letter \textquote{E} as \textquote{a rectangle with rounded corners defined by twelve lines}, which leads to an inaccurate generation result.
        Our annotation is consistent with the model geometry and leads to a correct result.
    }
    \label{fig:annot_comparison_3}
\end{figure*}

\Crefrange{fig:annot_comparison_1}{fig:annot_comparison_3} provide qualitative comparisons between our annotations and the expert Text2CAD annotations.
For reference, we also show CAD designs generated by models trained on each type of annotation (both models predict FeatureScript code).
Text2CAD annotations often omit parts of the geometry (\eg, in \cref{fig:annot_comparison_1} only a subset of the sketch is extruded) or describe features imprecisely (\cref{fig:annot_comparison_2,fig:annot_comparison_3}), which leads to incomplete or inaccurate generation results.
Our representation-aligned descriptions support more accurate and complete CAD model generation.

\section{Evaluation details}
\label{sec:sup_eval}

In this section, we provide details of the quantitative evaluation of generated CAD designs.
Unless otherwise stated, we compare generated and reference shapes by first sampling point clouds on their boundary surfaces.
Let $X = \{x_i\}_{i=1}^{|X|}$ and $Y = \{y_j\}_{j=1}^{|Y|}$ be point sets sampled from the generated and reference meshes, respectively, with $x_i, y_j \in \mathbb{R}^3$.

\paragraph{Chamfer Distance (CD)} measures the geometric discrepancy between the generated and reference 3D models.
It is defined as the symmetric average squared distance from each point in one cloud to its nearest neighbor in the other:
\begin{equation}
\label{eq:cd}
\begin{split}
d_{\mathrm{CD}}(X,Y)
= &~\frac{1}{|X|}\sum_{x \in X} \min_{y \in Y} \lVert x - y \rVert_2^2 \\
  &+ \frac{1}{|Y|}\sum_{y \in Y} \min_{x \in X} \lVert x - y \rVert_2^2.
\end{split}
\end{equation}
Chamfer Distance simultaneously captures how well the generated shape covers the reference surface (recall) and how close it stays to it (precision).
We report the median Chamfer Distance across the dataset relative to the size of the CAD model, scaled by $10^3$ for convenience.

\paragraph{Edge Chamfer Distance (ECD)} computes the Chamfer Distance according to \cref{eq:cd}, but restricts it to points $X^{E} \subset X$ and $Y^{E} \subset Y$ near the edges of the generated and reference meshes.
It assesses the fidelity of sharp geometric features which are important in industrial design.
To detect edge points, we use a local vicinity test.
For each point, we query all neighbors within a radius of $r = 0.004$ (in normalized unit-scale space) using a ball query over the point cloud.
A point is classified as an edge if \emph{any} neighbor within this vicinity exhibits a sufficiently different normal, i.e., when the absolute dot product satisfies $\lvert n_i^\top n_j \rvert < 0.2$.
We report the median Edge Chamfer Distance across the dataset relative to the size of the CAD model, scaled by $10^3$.

\paragraph{Normal Consistency (NC)} evaluates the consistency of the orientations of the generated and reference 3D surfaces.
For each $x \in X$ we denote by $y_x \in Y$ its nearest neighbor, and similarly $x_y$ for $y \in Y$.
Let $n_x$ and $n_y$ be unit normals at points $x$ and $y$.
Normal Consistency is then defined as
\begin{equation}
\label{eq:nc}
\begin{split}
\mathrm{NC}(X,Y)
= \tfrac{1}{2} \Big(
&\tfrac{1}{|X|}\sum_{x \in X} n_x \cdot n_{y_x} \\
+&\tfrac{1}{|Y|}\sum_{y \in Y} n_y \cdot n_{x_y}
\Big),
\end{split}
\end{equation}
where the averages are taken over the sampled points.
Values close to $1$ indicate that the corresponding surfaces are oriented consistently.
We report the median Normal Consistency across the dataset.

To compute CD, ECD, and NC, we sample 100k points from the reference and generated point clouds.

\paragraph{Coverage (COV)} assesses how well the set of generated shapes~$G$ covers the set of reference shapes~$S$.
For each $X \in G$, we denote by $\mathrm{NN}_S(X)$ its nearest neighbor in $S$ according to $d_{\mathrm{CD}}$.
The Coverage is the fraction of reference shapes that are matched at least once:
\begin{equation}
\label{eq:cov}
\mathrm{COV}(S,G)
=
\frac{1}{|S|}
\Big|
\{\mathrm{NN}_S(X) : X \in G\}
\Big|.
\end{equation}
Higher Coverage indicates that generated samples cover a larger portion of the reference shape space.
We report Coverage as a percentage.

\paragraph{Minimal Matching Distance (MMD)} measures how well the distribution of generated shapes approximates the reference distribution.
For each reference shape $Y \in S$ we compute the Chamfer Distance to its nearest generated neighbor in $G$ and average:
\begin{equation}
\mathrm{MMD}(S, G)
=
\frac{1}{|S|}
\sum_{Y \in S}
\min_{X \in G} d_{\mathrm{CD}}(X, Y).
\label{eq:mmd}
\end{equation}
Lower Minimal Matching Distance means that, on average, every reference shape is well approximated by some generated shape.
We report Minimal Matching Distance as the mean squared Euclidean distance on unit-normalized shapes scaled by $10^3$.

\paragraph{Jensen-Shannon Divergence (JSD)} is a statistical measure of similarity between probability distributions.
Here, it quantifies how similar the spatial point distributions of the reference shapes $S$ and the generated shapes $G$ are.
To compute Jensen-Shannon Divergence, the 3D space is discretized into a regular voxel grid, and each point in the sets is assigned to an $i$-th voxel, yielding empirical distributions $P_S$ and $P_G$ over voxels.
The Jensen-Shannon Divergence is then calculated as
\begin{equation}
\label{eq:jsd}
\begin{split}
\mathrm{JSD}(P_S,P_G)
= \tfrac{1}{2} &\, D_{\mathrm{KL}}(P_S \| M) \\
             + \tfrac{1}{2} &\, D_{\mathrm{KL}}(P_G \| M),
\end{split}
\end{equation}
where $M = \tfrac{1}{2}(P_S + P_G)$ and
\begin{equation}
\label{eq:kl}
D_{\mathrm{KL}}(P\|Q)
=
\sum_i P(i)\,\log\frac{P(i)}{Q(i)}.
\end{equation}
Smaller Jensen-Shannon Divergence indicates closer agreement between the distributions of reference and generated geometry.
We report Jensen-Shannon Divergence scaled by $10^2$.

Following DeepCAD~\cite{wu2021deepcad}, we randomly sample 3k shapes from each of the reference and generated sets, repeat this evaluation process 10 times, and report average scores for the COV, MMD, and JSD metrics. We sample 2k points from the reference and generated point clouds to compute these three metrics.

\paragraph{Invalidity Ratio (IR)} is the fraction of design histories that fail to construct into a valid solid (\eg, due to CAD kernel errors or invalid geometry or topology).
Let $N_{\mathrm{gen}}$ be the total number of generated sequences and $N_{\mathrm{inv}}$ the number of sequences failed to construct.
The Invalidity Ratio is defined as
\begin{equation}
\mathrm{IR}
=
\frac{N_{\mathrm{inv}}}{N_{\mathrm{gen}}} * 100\%.
\label{eq:ir}
\end{equation}
Lower Invalidity Ratio indicates that the model produces compilable CAD programs more reliably.
We report IR as a percentage.

    {\clearpage\newpage\small\bibliographystyle{ieeenat_fullname}\bibliography{src/bib}}
\end{document}